\documentclass[letterpaper]{article} 
\usepackage[preprint]{arxiv_style}
\usepackage[hyphens]{url}  
\usepackage{graphicx} 
\usepackage{natbib}  
\usepackage{caption} 
\usepackage{algorithm}
\usepackage{algorithmic}

\usepackage{newfloat}
\usepackage{listings}
\DeclareCaptionStyle{ruled}{labelfont=normalfont,labelsep=colon,strut=off} 
\floatstyle{ruled}
\newfloat{listing}{tb}{lst}{}
\floatname{listing}{Listing}

\usepackage{booktabs}

\usepackage{amsmath}
\usepackage{amssymb}
\usepackage{array}
\usepackage{multirow}

\title{Understanding Knowledge Transfer Mechanism in Heterogeneous MLLM Fusion: A Simple Linear Approach}
\author{
    Yinghao Hou,
    Jiahe Fan,
    Yuanhao Pu,
    Zongyuan Chen,
    Hong Xie\corresponding
}
\affiliations{
    University of Science and Technology of China\\
    Hefei, Anhui, China\\
    yhhou@mail.ustc.edu.cn, xiehong2018@foxmail.com
}

\begin{document}

\maketitle

\begin{abstract}
Training-free fusion of heterogeneous multimodal large language models (MLLMs) provides a direct route for cross-scale capability transfer, yet improvements in aggregate performance do not reveal what a smaller model actually inherits. Existing studies are largely designed and evaluated on limited task sets or aggregate metrics; as evaluation expands to broader task collections, whether different capabilities can transfer across scales remains poorly understood. To investigate this question, we introduce Cross-Scale Directional Parameter Injection (CDPI), a simple linear probe to analyze cross-scale knowledge transfer during heterogeneous fusion. A local theoretical analysis indicates that knowledge transfer selectivity
is determined at first order by capability-dependent responses to a shared
injection direction, while second-order curvature effects constrain the
effective transfer regime. Across four Qwen3-VL model pairs and twelve multimodal benchmarks, our experiments reveal a consistent pattern of selectivity: gains concentrate on reasoning, particularly high-level reasoning, whereas perception performance remains close to that of the original target model. Component-wise ablations further show that high-level reasoning gains arise primarily from the language model, while ratio analysis finds that positive selective transfer occurs mainly in the small-ratio regime. 
These findings recast cross-scale heterogeneous MLLM fusion as selective
language-side reasoning transfer within a narrow, low-interference regime,
rather than broad capability inheritance.
\end{abstract}


\section{Introduction}

Heterogeneous model fusion combines differently structured models in parameter
space for training-free knowledge transfer
\cite{cui2026transport,du2025adamms}.
We study cross-scale fusion within one multimodal large language model (MLLM)
family~\cite{alayrac2022flamingo,li2023blip2,liu2023visual}, asking what
transfers, where it resides, and how layer mapping shapes the transfer.

Model fusion largely assumes identical architectures, using weight averaging,
task vectors, or conflict resolution to integrate capabilities across tasks and
domains
\cite{wortsman2022modelsoups,ilharco2023taskarithmetic,yadav2023ties,yu2024supermario}.
Multimodal fusion generally shares this assumption
\cite{qu2025uqmerge,zeng2025robustmerge}. Recent heterogeneous methods instead
bridge differences in width, depth, or representation space through learned
mappings, latent representations, activation statistics, or optimal transport
\cite{soro2026lsmerge,cui2026transport,du2025adamms,wei2026optmerge}.
Yet their aggregate- or task-level evaluations reveal little about which
capabilities transfer or which parameters and components carry them, leaving
heterogeneous MLLM transfer mechanisms unclear.

\begin{figure*}[t]
    \centering
    \includegraphics[width=\textwidth]
    {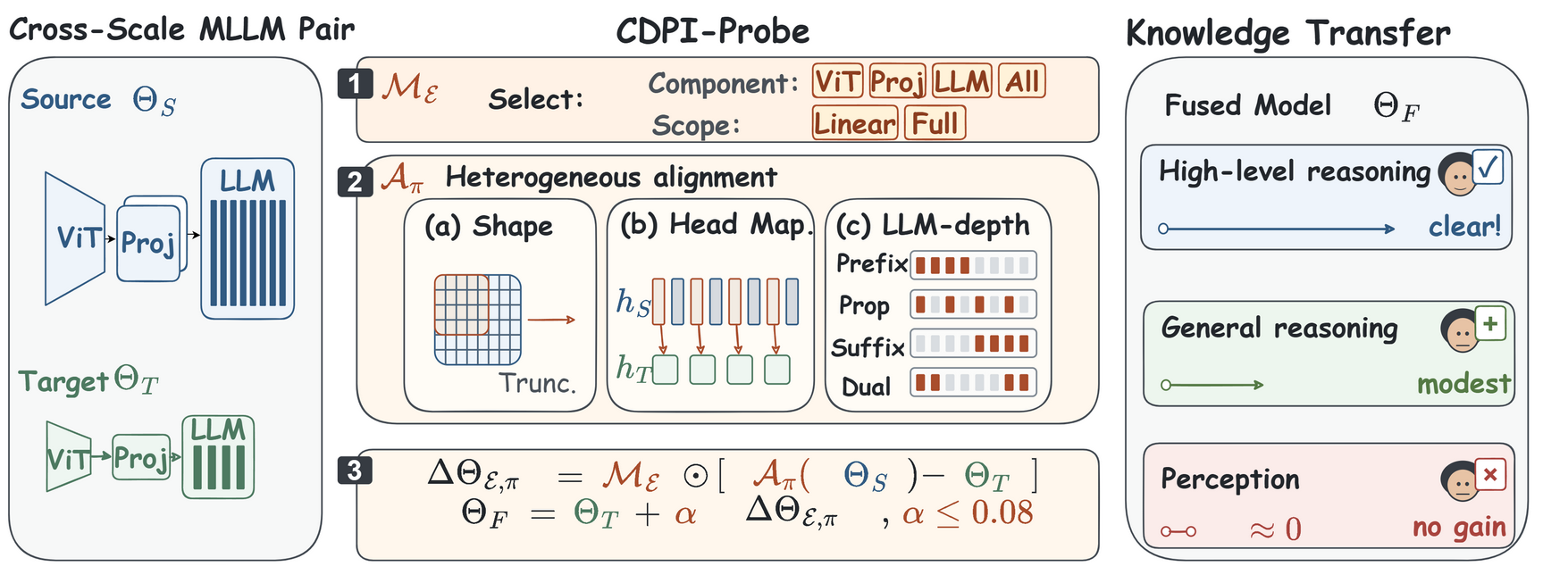}
    \caption{Overview of CDPI. A larger source MLLM is deterministically
    aligned with a smaller target; parameter scope, component, layer mapping,
    and injection ratio define the projected source direction. Resulting
    capability changes reveal the transferred knowledge and its determinants.}
    \label{fig:capability_probe}
\end{figure*}

Studying this mechanism raises three challenges: architectural differences
preclude element-wise parameter correspondence; transfer signals may reside in
different parameter subsets or MLLM components; and capability changes can
conflate transfer with interference from truncation, misalignment, and component
mismatch. We therefore introduce Cross-Scale Directional Parameter Injection
(CDPI; Figure~\ref{fig:capability_probe}), a linear probe that projects source
parameters into the target space through shape truncation, head mapping, and
layer mapping, then moves the target slightly along the projected direction.
Varying parameter scope, component, layer mapping, and injection ratio reveals
where transferable knowledge resides and what shapes its transfer.

A local second-order analysis attributes selective transfer to competition
between first-order signal and second-order curvature responses. Across four Qwen3-VL
pairs and twelve benchmarks, gains concentrate on high-level reasoning while
perception stays near the target. Linear weights and the LLM carry most gains; effective layer mappings are
pair-dependent, and transfer is confined to a narrow, low-interference ratio
regime.

Our main contributions are summarized as follows:
\begin{itemize}
    \item We develop a framework for analyzing knowledge
    transfer in cross-scale heterogeneous MLLM fusion, using CDPI as a
    training-free probe that injects a projected source direction into the
    target. A local quadratic analysis further decomposes capability gain
    into a first-order transfer signal and a second-order curvature response.
    \item We show that cross-scale knowledge transfer is
    capability-selective: gains concentrate in reasoning, especially
    high-level reasoning, whereas perception remains
    unchanged.
    \item We characterize how parameter selection, model component,
    layer mapping, and injection ratio shape knowledge transfer.
    Linear-layer weights generally yield higher transfer efficiency,
    high-level reasoning gains arise primarily from the LLM, effective
    layer mappings depend on the source--target pair, and positive
    selective transfer occurs mainly at small injection ratios.
\end{itemize}

\section{Related Work}

\paragraph{Homogeneous Model Fusion.}
Homogeneous fusion integrates compatible checkpoints without joint training
through averaging, task arithmetic, conflict resolution, sparsification,
adaptive weighting, or representation alignment
\cite{yang2024adamerging,yang2024surgery,
 wortsman2022modelsoups,ilharco2023taskarithmetic,yadav2023ties,
 yu2024supermario,matena2022fisher,huang2024emr}.
These methods generally assume identical architectures and element-wise
parameter correspondence across the models being fused. In the multimodal
setting, UQ-Merge and RobustMerge study task-specialized MLLMs but likewise
focus on compatible parameter structures
\cite{qu2025uqmerge,zeng2025robustmerge}. Alignment-based methods instead
establish cross-model correspondence through permutation or feature matching
\cite{ainsworth2023gitrebasin,jordan2023repair,stoica2024zipit,
stoica2025knots}.

\paragraph{Heterogeneous Model Fusion.}
Heterogeneous fusion relaxes structural compatibility by aligning models that
differ in width, depth, or representation space. LS-Merge uses latent
representations, Transport and Merge uses activation statistics and optimal
transport, and recent work extends alignment to multimodal models
\cite{imfeld2024transformer,soro2026lsmerge,cui2026transport,
du2025adamms,wei2026optmerge}.
These studies demonstrate that structurally heterogeneous models can be fused
through learned mappings or representation alignment. However, aggregate- or
task-level evaluations leave capability-wise transfer underexplored.

\paragraph{Knowledge Transfer and Capability Analysis.}
Knowledge distillation and fusion transfer model knowledge through output
distributions, intermediate representations, or modular updates across both
LLMs and MLLMs
\cite{hinton2015distilling,gu2024minillm,cai2025llavakd,
wan2024fusellm,wan2025fusechat,du2026graftllm}.
The resulting student capabilities are jointly shaped by the transfer data,
distillation objective, and gradient-based optimization. Model stitching
instead learns mappings between intermediate representations to analyze
functional compatibility across models \cite{bansal2021stitching}. These
studies examine the transferability of model knowledge and functionality from
the perspective of output behavior or intermediate representations, but
provide limited understanding of capability-level changes induced directly by
heterogeneous parameter fusion.

\section{Method}

\subsection{Cross-Scale Directional Parameter Injection}

Given a larger source model $\Theta_S$ and a smaller target model
$\Theta_T$ from the same MLLM family~\cite{bai2025qwen3vl} (e.g.,
Qwen3-VL-8B-Instruct and Qwen3-VL-2B-Instruct), CDPI injects a projected
source direction into the target parameter space. Differences in hidden
size, attention-head count, and language-model depth preclude direct
element-wise fusion. We partition each model into three components: the
vision encoder (ViT), cross-modal projector (Proj), and language model (LLM):
\begin{equation}
\Theta_X =
\left(
\Theta_X^{\mathrm{ViT}},
\Theta_X^{\mathrm{Proj}},
\Theta_X^{\mathrm{LLM}}
\right),
\quad X \in \{S,T\}.
\end{equation}

Denoting the resulting parameters by $\Theta_F$, the training-free injection
is formulated as
\begin{equation}
\Theta_F
=
\Theta_T
+
\alpha \Delta\Theta_{\mathcal E,\pi},
\end{equation}
where $\Delta\Theta_{\mathcal E,\pi}$ is the projected source
direction injected into the target model; $\mathcal E$ is the
injection-space configuration that specifies the selected model
components and parameter scope; $\pi$ specifies the LLM layer mapping; and
$\alpha\in[0,1]$ is the injection ratio that controls the distance
traveled along this direction.

We define the injection space as $\mathcal E=(C,P)$, where
$C\subseteq\{\mathrm{ViT},\mathrm{Proj},\mathrm{LLM}\}$ specifies the
selected components. When all three components are selected, i.e.,
$C=\{\mathrm{ViT},\mathrm{Proj},\mathrm{LLM}\}$, we abbreviate this
configuration as $C=\mathrm{All}$. The parameter scope
$P\in\{\texttt{linear},\texttt{full}\}$ restricts the injection to
linear-layer parameters or includes all alignable parameters within
the selected components, respectively. For example,
$\mathcal E=(\{\mathrm{ViT}\},\texttt{linear})$ injects only the
linear-layer parameters of the ViT. Detailed parameter-selection
rules and exclusions are provided in the supplementary material.

Because the source and target parameters have different shapes, we define a heterogeneous alignment operator $\mathcal A_\pi$ that maps $\Theta_S$ into the target parameter space, making it shape-compatible with $\Theta_T$.

Let $\mathcal M_{\mathcal E}$ be the binary mask associated with $\mathcal E$, with ones at the selected positions and zeros elsewhere. The projected source direction injected into the target model is defined as
\begin{equation}
\Delta\Theta_{\mathcal E,\pi}
=
\mathcal M_{\mathcal E}
\odot
\left[
\mathcal A_\pi(\Theta_S)-\Theta_T
\right],
\end{equation}
where $\odot$ denotes element-wise or block-wise multiplication. The final fused parameters are therefore
\begin{equation}
\Theta_F
=
\Theta_T
+
\alpha \mathcal M_{\mathcal E}
\odot
\left[
\mathcal A_\pi(\Theta_S)-\Theta_T
\right].
\end{equation}

\subsection{A Theoretical Interpretation of CDPI}

For notational simplicity, let $\Delta \triangleq \Delta\Theta_{\mathcal E,\pi}$. For a capability category $c$, let $\mathcal L_c(\Theta)$ denote its generalization loss. We define the local capability gain as 
\begin{equation}
\operatorname{Gain}_c=-(\mathcal L_c(\Theta_F)-\mathcal L_c(\Theta_T))
\end{equation}

A second-order expansion around $\Theta_T$ gives
\begin{equation}
\mathcal L_c(\Theta_F)-\mathcal L_c(\Theta_T)
\approx
\alpha\nabla_\Theta\mathcal L_c(\Theta_T)^\top\Delta
+
\frac{\alpha^2}{2}\Delta^\top H_c\Delta
\end{equation}
where $H_c=\nabla_\Theta^2\mathcal L_c(\Theta_T)$. We define the first-order transfer signal as
$A_c=-\nabla_\Theta\mathcal L_c(\Theta_T)^\top\Delta$ and the second-order
curvature response as $I_c=\Delta^\top H_c\Delta$.
The local capability gain can therefore be approximated as
\begin{equation}
\operatorname{Gain}_c
\approx
\alpha A_c-\frac{\alpha^2}{2}I_c
\end{equation}

Here, $A_c$ measures whether the projected source direction is aligned
with the local improvement direction of capability $c$, while $I_c$
captures the curvature response along this direction. When $I_c>0$,
we interpret the resulting second-order penalty as effective structural
interference, which may reflect shape truncation, parameter
misalignment, or component mismatch. In particular, if $A_c>0$, the
projected source direction contains a first-order transfer signal that
benefits capability $c$. When both $A_c>0$ and $I_c>0$, the locally
optimal ratio under the quadratic approximation is
$\alpha_c^\star=A_c/I_c$, and the gain remains positive when
$0<\alpha<2A_c/I_c$.

This explains why small-ratio injection can be effective. When $I_c>0$, the first-order transfer signal grows linearly with $\alpha$, whereas the second-order curvature penalty grows quadratically. A small $\alpha$ preserves the positive signal while limiting this penalty, but larger $\alpha$ lets the curvature term offset the first-order benefit and degrade performance.

\paragraph{Selective Capability Transfer.}
When $\alpha$ is small, the local gain is dominated by the first-order term, i.e., $\operatorname{Gain}_c\approx-\alpha\nabla_\Theta\mathcal L_c(\Theta_T)^\top\Delta$. Let $\phi_c=\angle(\Delta,-\nabla_\Theta\mathcal L_c(\Theta_T))$ denote the angle between the projected source direction and the local improvement direction of capability $c$. Then
\begin{equation}
\operatorname{Gain}_c
\approx
\alpha
\left\|\nabla_\Theta\mathcal L_c(\Theta_T)\right\|_2
\left\|\Delta\right\|_2
\cos\phi_c
\end{equation}

Thus, the first-order gain depends on the target model's sensitivity
to capability $c$, the injection ratio, and the directional alignment.
For a fixed projected source direction $\Delta$, different
capabilities generally have different local improvement directions.
Therefore, for two capabilities $c_1$ and $c_2$, $\phi_{c_1}\neq\phi_{c_2}$, leading to selective capability gains. For example, if $\cos\phi_{\mathrm{reason}}>0$ while $\cos\phi_{\mathrm{perception}}\approx0$, reasoning capability improves while perception remains nearly unchanged. This indicates that a projected source direction does not improve all
capabilities uniformly.

\subsection{Heterogeneous Alignment Operator $\mathcal{A}_{\pi}$}

The heterogeneous alignment operator $\mathcal A_\pi$ maps the source
parameters into the target parameter space while accounting for
differences in tensor shape, attention-head configuration, and
language-model depth. It consists of three operations: parameter
correspondence and shape truncation, structure-preserving
attention-head mapping, and language-model layer mapping.

\paragraph{Parameter Correspondence and Shape Truncation.}
For each target parameter, we first identify its source counterpart based on the model component, parameter name, and tensor type. For repeated Transformer layers in the LLM, the source-layer index is determined by the layer mapping $\pi$, while the internal parameter path remains unchanged.

Given a source tensor $X\in\mathbb{R}^{d_1^S\times\cdots\times d_m^S}$ and a target shape $s=(d_1^T,\ldots,d_m^T)$, where $d_r^S\geq d_r^T$, we retain the leading slice of the source tensor along each mismatched dimension:
\begin{equation}
\operatorname{Trunc}(X;s)
=
X[0:d_1^T,\ldots,0:d_m^T].
\label{eq:shape_truncation}
\end{equation}
The truncated tensor has the same shape as the corresponding target parameter.

\paragraph{Attention-Head Mapping.}
When the source and target models have different numbers of query heads, directly truncating the flattened attention projection may split complete head blocks. We therefore select complete source query-head blocks at approximately uniform intervals. Let $H_S^q$ and $H_T^q$ denote the numbers of query heads in the source and target models, respectively. The $j$-th target query head is mapped to the following source head:
\begin{equation}
\rho_q(j)
=
\left\lfloor
j\frac{H_S^q}{H_T^q}
\right\rfloor,
\qquad
j=0,\ldots,H_T^q-1.
\label{eq:query_head_mapping}
\end{equation}
After selecting the query-head blocks, we truncate the corresponding projection tensors to the target shapes. The source--target model pairs considered in this work have the same number of key/value heads and therefore require no additional K/V head mapping.

\begin{table*}[!t]
\centering
\normalsize
\setlength{\tabcolsep}{5pt}
\begin{tabular}{
  @{}
  >{\raggedright\arraybackslash}m{0.16\textwidth}
  >{\raggedright\arraybackslash}m{0.22\textwidth}
  >{\raggedright\arraybackslash}m{0.54\textwidth}
  @{}
}
\hline
Capability & Subcategory & Benchmarks \\
\hline

\multirow[c]{2}{*}{Reasoning}
& High-level Reasoning
& MMMU-Pro, MathVista, MATH-Vision, VisuLogic, VisualPuzzles
  \cite{yue2025mmmupro,lu2024mathvista,wang2024mathvision,
  xu2026visulogic,song2025visualpuzzles} \\

& General Reasoning
& MMMU, MMVU
  \cite{yue2024mmmu,zhao2025mmvu} \\

\hline

Perception
& Visual Perception
& MME, MMStar, BLINK, OCRBench, ChartQA
  \cite{fu2025mme,chen2024mmstar,fu2024blink,liu2024ocrbench,
  masry2022chartqa} \\

\hline
\end{tabular}

\caption{Capability-oriented benchmark taxonomy.}
\label{tab:benchmark_taxonomy}
\end{table*}

\paragraph{Language-Model Layer Mapping.}
Let $L_S$ and $L_T$ denote the numbers of source and target LLM layers, respectively, with $L_S\geq L_T$. The mapping
$\pi:\{0,\ldots,L_T-1\}\rightarrow\{0,\ldots,L_S-1\}$
maps the $i$-th target layer to the $\pi(i)$-th source layer. We consider four mapping strategies (Prefix, Depth-proportional, Suffix, and Dual-end):
\begin{equation}
\begin{aligned}
\pi_{\mathrm{pre}}(i)
&=i,\\
\pi_{\mathrm{prop}}(i)
&=
\left\lfloor
i\frac{L_S}{L_T}
\right\rfloor,\\
\pi_{\mathrm{suf}}(i)
&=i+(L_S-L_T),\\
\pi_{\mathrm{dual}}(i)
&=
\begin{cases}
i,
& i<\lfloor L_T/2\rfloor,\\
L_S-L_T+i,
& i\geq\lfloor L_T/2\rfloor.
\end{cases}
\end{aligned}
\label{eq:layer_mapping}
\end{equation}
Prefix and Suffix select the early and late source layers, respectively; Depth-proportional samples source layers approximately uniformly along model depth; and Dual-end retains information from both shallow and deep source layers. After layer mapping, we replace only the layer index in the parameter path, while keeping the internal parameter path within each Transformer block unchanged.

Together, these operations define $\mathcal A_\pi(\Theta_S)$, whose
parameter structure and tensor shapes are compatible with those of
the target model $\Theta_T$.

\begin{table}[!t]
\centering
\normalsize
\setlength{\tabcolsep}{4pt}
\begin{tabular}{
  @{}
  >{\raggedright\arraybackslash}m{0.32\linewidth}
  >{\raggedright\arraybackslash}m{0.62\linewidth}
  @{}
}
\hline
Factor & Values \\
\hline

Component $C$
& ViT, Proj, LLM, \textbf{All} \\

Parameter scope $P$
& \textbf{\texttt{linear}}, \texttt{full} \\

Layer mapping $\pi$
& \textbf{Prefix}, Depth-proportional, Suffix, Dual-end \\

\hline
\end{tabular}
\caption{CDPI configurations. Default settings are in \textbf{bold}.}
\label{tab:fusion_configurations}
\end{table}

\section{Experiments}
\label{sec:experiments}

We examine what transfers, which parameter scope and component carry it,
how layer mapping routes it, and how transfer changes with the injection ratio.

\subsection{Experimental Setup}

\paragraph{Models.}
We use Qwen3-VL-Instruct models~\cite{bai2025qwen3vl} at four scales
(2B, 4B, 8B, and 32B)
and evaluate four source--target pairs: 4B$\rightarrow$2B,
8B$\rightarrow$2B, 32B$\rightarrow$4B, and 32B$\rightarrow$8B.

\paragraph{Capability Groups.}
We organize the evaluation benchmarks into three capability
groups~\cite{li2024seedbench,yu2024mmvet}:
high-level reasoning, general reasoning, and perception. High-level reasoning
covers multi-step visual reasoning, mathematical reasoning, logical inference,
and compositional problem solving. General reasoning evaluates multimodal
reasoning over diverse visual and knowledge-intensive content. Perception
primarily evaluates visual recognition, OCR, and chart understanding. We group
benchmarks by their dominant evaluation demands rather than treating these
categories as mutually exclusive. Perception-oriented benchmarks place greater
emphasis on direct visual understanding than the two reasoning groups, although
some also require reasoning.
Table~\ref{tab:benchmark_taxonomy} summarizes the benchmark taxonomy.

\paragraph{Transfer Metric.}
For a benchmark group $\mathcal B$, we measure empirical capability transfer
by the average score gain over the target model:
\begin{equation}
\operatorname{Gain}(\mathcal B)
=
\frac{1}{|\mathcal B|}
\sum_{b\in\mathcal B}
\left(
S_F(b)-S_T(b)
\right),
\label{eq:empirical_gain}
\end{equation}
where $S_F(b)$ and $S_T(b)$ denote the fused-model and target-model scores on
benchmark $b$, respectively. A positive value indicates improvement over the
original target model.

\paragraph{Experimental Protocol.}
We evaluate configurations of $(C,P,\pi)$ by varying one factor at a
time. Unless otherwise specified, we use $C=\mathrm{All}$,
$P=\texttt{linear}$, and Prefix mapping. To characterize the observed
boundary of positive transfer within the local regime, for each benchmark we
evaluate $\alpha\in\{0.01,0.02,0.04,0.06,0.08\}$ and report the result
at the ratio where positive transfer peaks, treating this as an upper-bound estimate of CDPI's beneficial effect. For detailed ratio results, Finding~1 further examines a representative source--target pair ($32\mathrm{B}\rightarrow4\mathrm{B}$), reporting results across all injection ratios for individual checkpoint characterization.
Complete ratio-wise results for all model pairs, together
with additional implementation details, are provided in the
supplementary material. All evaluations use
\texttt{lmms-eval}~\cite{zhang2025lmmseval} with benchmark-standard
generation and scoring settings and no GPT-based judging.

\begin{figure*}[!t]
\centering
\includegraphics[width=\textwidth]
{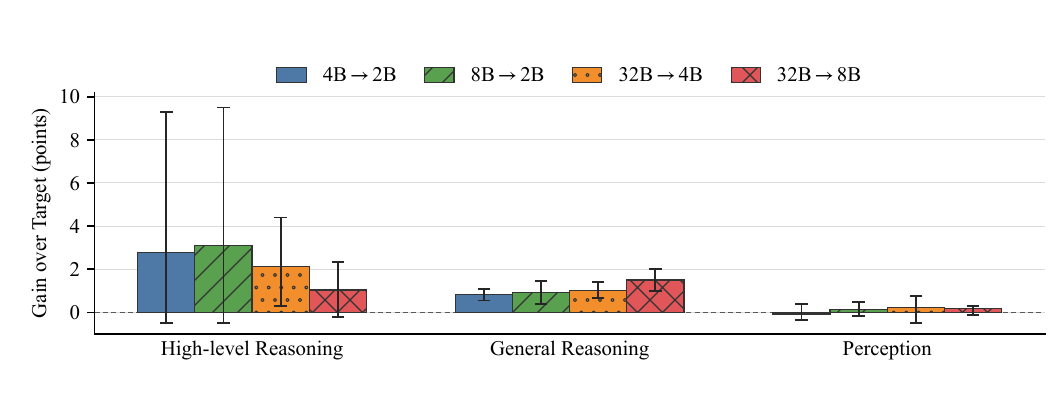}
\caption{Mean gains over the target model across capability groups;
error bars show the benchmark range. Cross-scale transfer is concentrated
in reasoning, particularly high-level reasoning.}
\label{fig:capability_gain}
\end{figure*}

\begin{table*}[!t]
\centering
\normalsize
\setlength{\tabcolsep}{4.2pt}

\begin{tabular}{@{}l|ccccc|ccc@{}}
\hline
& \multicolumn{5}{|c|}{High-level reasoning benchmarks}
& \multicolumn{3}{c}{Capability average} \\
\cline{2-6}\cline{7-9}

Setting
& MMMU-Pro
& MathVista
& MATH-Vision
& VisuLogic
& Visual Puzzles
& High-level
& General
& Perception \\
\hline

Source (32B)
& 45.61 & 61.50 & 33.22 & 24.80 & 33.39
& 39.70 & 62.59 & 79.21 \\

Target (4B)
& 31.79 & 53.30 & 19.41 & 26.80 & 29.79
& 32.22 & 51.59 & 75.12 \\
\hline

$\alpha=0.01$
& \textbf{33.74} & 54.60 & 19.06 & 26.30 & 29.02
& 32.54 & 51.77 & \textbf{75.24} \\

$\alpha=0.02$
& 33.24 & \textbf{55.30} & \textbf{23.81} & 26.40 & \textbf{31.74}
& \textbf{34.10} & \textbf{52.45} & 74.96 \\

$\alpha=0.04$
& 32.90 & 54.10 & 22.64 & \textbf{27.10} & 30.48
& 33.44 & 52.22 & 73.65 \\

$\alpha=0.06$
& 31.62 & 51.10 & 21.05 & 26.20 & 30.05
& 32.00 & 51.17 & 72.53 \\

$\alpha=0.08$
& 28.96 & 50.10 & 20.07 & 26.70 & 31.68
& 31.50 & 49.74 & 68.94 \\
\hline
\end{tabular}

\caption{Ratio-wise results for the representative
32B$\rightarrow$4B pair. We report individual benchmarks for high-level
reasoning and capability-level averages for all three groups. Each row
applies a single common injection ratio across all tasks.}
\label{tab:benchmark_alpha_32b4b}
\end{table*}

\subsection{Key Findings}

\subsubsection{Finding 1: Cross-Scale Knowledge Transfer Is Concentrated in Reasoning Capabilities}

We first examine which capabilities are transferred through CDPI.
Figure~\ref{fig:capability_gain} reports capability-level gains across the
four source--target pairs. High-level reasoning receives the largest mean
gain for three of the four pairs, whereas perception remains within $0.24$
points of the target across all pairs. For the
32B$\rightarrow$4B results, the corresponding gains are $+2.12$, $+1.03$,
and $+0.24$ for high-level reasoning, general reasoning, and perception.

To examine this pattern at both the task and fixed-ratio levels,
Table~\ref{tab:benchmark_alpha_32b4b} reports detailed high-level reasoning
results for 32B$\rightarrow$4B, together with averages for all three
capability groups. At the fixed ratio
$\alpha=0.02$, four of the five high-level benchmarks outperform the target;
MATH-Vision increases from $19.41$ to $23.81$ ($+4.40$). At $\alpha=0.04$, all five
high-level benchmarks improve. Importantly, this selectivity is not an
artifact of task-wise ratio selection: at a single fixed ratio of
$\alpha=0.02$, the 32B$\rightarrow$4B pair improves high-level and general
reasoning by $1.88$ and $0.86$ points, respectively, while perception changes
by only $-0.16$ points. Corresponding fixed-ratio results for the other
source--target pairs are provided in the supplementary material.

Together, the task-level and ratio-wise results show that cross-scale
transfer primarily benefits reasoning, especially high-level tasks involving
multi-step, mathematical, and compositional reasoning.

\subsubsection{Finding 2: Parameter Scope Shapes Transfer Efficiency}

We fix the remaining factors and vary only the parameter scope $P$,
comparing \texttt{All-linear} and \texttt{All-full} injection across the
four source--target pairs.

\begin{figure*}[!t]
\centering
\includegraphics[width=\linewidth]
{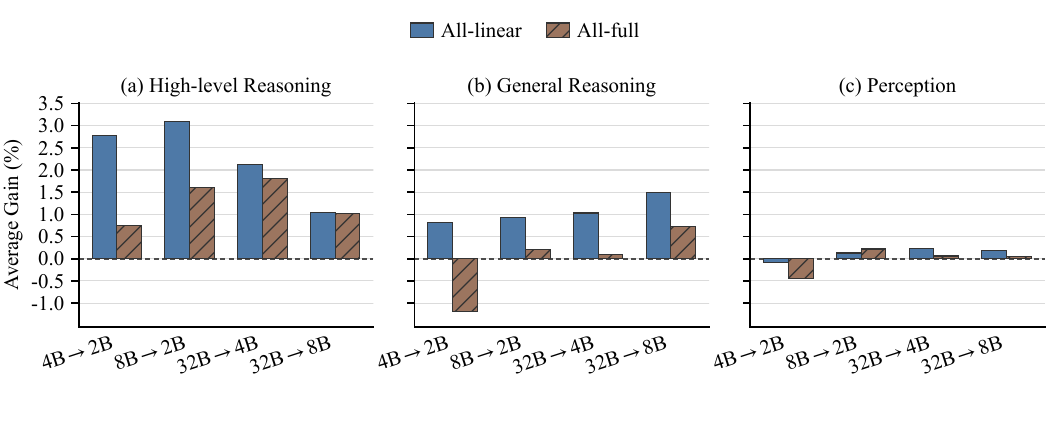}
\caption{Capability gains under \texttt{All-linear} and
\texttt{All-full} injection. Restricting injection to linear layers
generally yields larger reasoning gains, while perception changes only
marginally.}
\label{fig:space_transfer}
\end{figure*}

The parameter scope substantially affects transfer efficiency. Averaged
across the four model pairs, \texttt{All-linear} improves high-level
reasoning by $+2.26$, compared with $+1.30$ for \texttt{All-full}.
For general reasoning, the corresponding gains are $+1.07$ and
approximately zero, while perception remains close to the target under
both settings. The 32B$\rightarrow$4B pair follows the same
pattern: \texttt{All-linear} yields gains of $+2.12$ and $+1.03$ for
high-level and general reasoning, compared with $+1.81$ and $+0.09$ for
\texttt{All-full}.

Thus, the parameter scope $P$ does not change the concentration of
knowledge transfer in reasoning, but restricting injection to linear-layer
weights generally transfers that knowledge more efficiently.

\subsubsection{Finding 3: High-Level Reasoning Transfer Primarily Originates from the LLM}

Having established that cross-scale knowledge transfer is concentrated in
high-level reasoning, we next identify which model component provides this
reasoning knowledge. We separately inject the ViT, Proj, and LLM components
and compare their high-level reasoning gains.

\begin{figure}[!t]
\centering
\includegraphics[width=\linewidth]{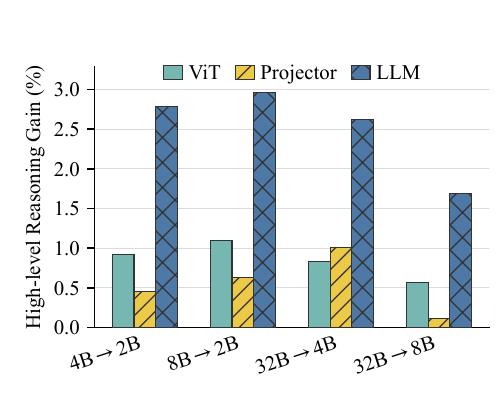}
\caption{Contributions of different model components to high-level
reasoning transfer. LLM injection produces the largest gain across all
four source--target pairs.}
\label{fig:module_source}
\end{figure}

LLM injection yields the largest high-level reasoning gain for every
source--target pair. The gains are $+2.78$, $+2.96$, $+2.62$, and $+1.69$ for
4B$\rightarrow$2B, 8B$\rightarrow$2B, 32B$\rightarrow$4B, and
32B$\rightarrow$8B, respectively, whereas ViT and Proj injections
produce substantially smaller gains.

These results indicate that cross-scale high-level reasoning transfer
primarily originates from the language model. Although the tasks
involve visual inputs, the gains are more strongly associated with the projected
source direction in the LLM component, carrying signals for
multi-step, logical, and compositional reasoning.

\subsubsection{Finding 4: Layer Mapping Shapes How Reasoning Enters the Target LLM}

Having identified the LLM as the primary source of high-level reasoning
transfer, we fix the injected component to the LLM and examine how the layer
mapping $\pi$ affects the path through which reasoning enters the target
model. We compare Prefix, Depth-proportional, Suffix, and Dual-end mappings.

\begin{figure}[!t]
\centering
\includegraphics[width=\linewidth]{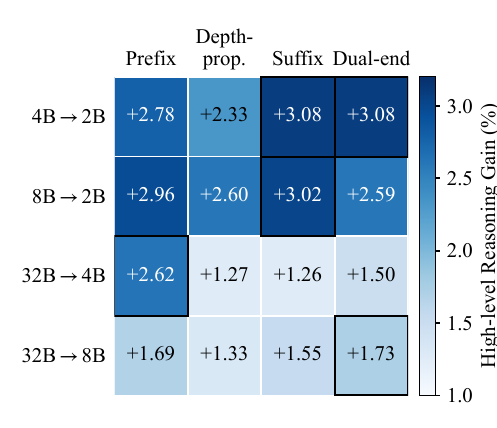}
\caption{High-level reasoning gains under four LLM layer mappings. The most
effective injection path depends on the source--target pair.}
\label{fig:alignment_heatmap}
\end{figure}

The best layer mapping differs across model pairs. For 4B$\rightarrow$2B,
Suffix and Dual-end achieve the largest gains. Suffix performs best for
8B$\rightarrow$2B, Prefix performs best for 32B$\rightarrow$4B, and Dual-end
achieves the largest gain for 32B$\rightarrow$8B.

No layer mapping is uniformly optimal. Edge-oriented mappings, including
Prefix, Suffix, and Dual-end, generally outperform Depth-proportional mapping,
but the most effective depth region depends on the source--target pair.
These results indicate that the high-level reasoning signal in the
projected source direction is not distributed uniformly across LLM
depth. Instead, the layer mapping $\pi$ determines how this signal
enters the target LLM.

\subsection{Analysis}

\subsubsection{Effect of the Injection Ratio on Capability Transfer}

We analyze how selective capability transfer changes with the
injection ratio $\alpha$. Fix $C=\mathrm{All}$,
$P=\texttt{linear}$, and Prefix layer mapping, and use
8B$\rightarrow$2B as the representative pair.
In addition to the main ratios
$\alpha\in\{0.01,0.02,0.04,0.06,0.08\}$, we include $\alpha=0.16$ and
$\alpha=0.32$ as stress tests. This experiment reports the capability gain at
each fixed ratio without selecting the best value.

\begin{figure}[!t]
\centering
\includegraphics[width=\linewidth]{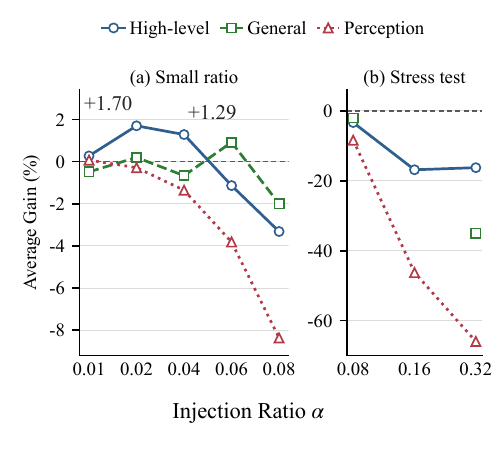}
\caption{Capability gains under different injection ratios for the
8B$\rightarrow$2B pair. Selective transfer emerges in the small-ratio regime,
whereas larger ratios cause substantial degradation. The general-reasoning
result at $\alpha=0.16$ is omitted due to an MMVU scoring artifact.}
\label{fig:ratio_curve}
\end{figure}

High-level reasoning reaches its largest gain at $\alpha=0.02$ and remains
positive at $\alpha=0.04$, while perception stays close to the target model or
begins to decline. At $\alpha=0.08$, both high-level reasoning and perception
fall below the target model, and the two larger stress-test ratios cause
substantially greater degradation.

Selective capability transfer therefore emerges primarily in the small-ratio
regime. This behavior is consistent with the interplay between the first-order
transfer signal and the second-order curvature response in our local analysis:
\[
\operatorname{Gain}_c
\approx
\alpha A_c-\frac{\alpha^2}{2}I_c.
\]
For $I_c>0$, the first-order transfer signal grows linearly with $\alpha$,
whereas the second-order curvature penalty grows quadratically. The latter
remains limited at small ratios but can increasingly offset the first-order
benefit as $\alpha$ increases, consistent with the observed degradation.

\subsubsection{Random Perturbations Do Not Reproduce Directional Transfer}

We examine whether the gains in the small-ratio regime can be reproduced by
generic parameter perturbations. Using the same 8B$\rightarrow$2B pair and
the same $C=\mathrm{All}$, $P=\texttt{linear}$ parameter subset, we compare
CDPI at $\alpha=0.02$ and $\alpha=0.04$ with bounded random multiplicative
perturbations evaluated over five seeds. The exact perturbation construction
is provided in the supplementary
material.

\begin{table}[!t]
\centering
\setlength{\tabcolsep}{1mm}
\begin{tabular}{@{}lcccc@{}}
\hline
Category
& Target(2B)
& $\alpha{=}0.02$
& $\alpha{=}0.04$
& Random \\
\hline
\multicolumn{5}{@{}l}{\textit{Capability averages}} \\

High-level
& 24.96
& \textbf{26.66}
& 26.25
& $24.02_{\pm 1.31}$ \\

General
& 46.44
& \textbf{46.64}
& 45.77
& $45.19_{\pm 0.37}$ \\

Perception
& \textbf{65.99}
& 65.70
& 64.63
& $64.91_{\pm 0.42}$ \\

\hline
\multicolumn{5}{@{}l}{\textit{High-level reasoning benchmarks}} \\

MMMU-Pro
& 23.76
& \textbf{24.57}
& 22.95
& $22.66_{\pm 2.73}$ \\

MathVista
& \textbf{48.80}
& 46.80
& 43.10
& $47.48_{\pm 0.56}$ \\

MATH-Vision
& 7.89
& \textbf{11.18}
& 9.87
& $9.34_{\pm 0.38}$ \\

VisuLogic
& 16.70
& 20.70
& \textbf{26.20}
& $13.72_{\pm 6.45}$ \\

VisualPuzzles
& 27.65
& \textbf{30.05}
& 29.11
& $26.88_{\pm 1.62}$ \\

\hline
\end{tabular}
\caption{CDPI versus random multiplicative perturbation on the
8B$\rightarrow$2B pair. Random results are averaged over five seeds,
with subscripts denoting standard deviations.}
\label{tab:random_perturbation}
\end{table}

The small-ratio transfer depends on the projected source direction rather than arbitrary parameter noise. As shown in Table~\ref{tab:random_perturbation}, CDPI at $\alpha=0.02$ lifts high-level reasoning from $24.96$ to $26.66$, whereas random perturbation yields only $24.02$ on average. The gap is larger at the benchmark level: MATH-Vision and VisualPuzzles gain $+3.29$ and $+2.40$, and VisuLogic improves by $+9.50$ at $\alpha=0.04$. Random perturbations capture none of these improvements and degrade all three capability averages.

\section{Conclusion}

We analyze cross-scale knowledge transfer in heterogeneous MLLM fusion
using CDPI, a training-free linear probe that injects a projected source
direction into the target parameter space. Across four Qwen3-VL model pairs
and twelve multimodal benchmarks, transfer gains concentrate on high-level
reasoning, while perception remains largely unchanged. Injecting linear-layer
weights is generally more effective; reasoning gains arise primarily from the
language model, and effective layer mappings vary across model pairs. Transfer
occurs mainly at small injection ratios, consistent with the interplay between
first-order transfer signals and second-order curvature responses in our local
analysis.
Taken together, these findings characterize cross-scale heterogeneous fusion as
a capability-selective knowledge-transfer process centered on language-side
reasoning within a narrow regime.

\setcounter{secnumdepth}{2}
\setcounter{equation}{0}
\setcounter{table}{0}
\setcounter{figure}{0}
\renewcommand{\theequation}{S\arabic{equation}}
\renewcommand{\thetable}{S\arabic{table}}
\renewcommand{\thefigure}{S\arabic{figure}}
\setcounter{dbltopnumber}{4}
\renewcommand{\dbltopfraction}{0.98}
\renewcommand{\dblfloatpagefraction}{0.70}
\setcounter{topnumber}{4}
\renewcommand{\topfraction}{0.98}
\renewcommand{\floatpagefraction}{0.70}

\appendix
\section*{Supplementary Material}

\section{Injection-Space Configuration and Mask Construction}
\label{sec:supp_injection_mask}

We provide additional details on the injection-space configuration and
parameter-selection mask used by Cross-Scale Directional Parameter Injection
(CDPI). The injection space is $\mathcal E=(C,P)$, where
$C\subseteq\{\mathrm{ViT},\mathrm{Proj},\mathrm{LLM}\}$ specifies the selected
model components and $P\in\{\texttt{linear},\texttt{full}\}$ specifies the
parameter scope. We use $C=\mathrm{All}$ as shorthand for selecting all three
components.

For $P=\texttt{linear}$, we select only the weight tensors of linear layers
within the components specified by $C$. These include the query, key, value,
and output projections in attention blocks, the linear matrices in
feed-forward networks, and the linear layers in the cross-modal projector,
where applicable. All other parameters remain unchanged.

For $P=\texttt{full}$, we select all alignable parameter tensors within the
specified components, including linear-layer weights and normalization-layer
parameters. A parameter is alignable if a corresponding source parameter can
be identified and mapped into the target shape by the heterogeneous alignment
operator $\mathcal A_{\pi}$. Parameters without a valid source counterpart are
not injected. Under both scopes, the target token-embedding layer and the LLM
output head remain unchanged.

Let $\theta_T^{(k)}$ denote the $k$-th target parameter tensor,
$\operatorname{comp}(k)$ its model component, and $\mathcal I_P$ the
parameter-index set specified by $P$. The active indices are
\begin{equation}
\mathcal K_{\mathcal E}
=
\left\{
k \,\middle|\,
\operatorname{comp}(k)\in C,\;
k\in\mathcal I_P
\right\}.
\label{eq:supp_active_parameter_set}
\end{equation}
Let $\mathbf{1}_k$ and $\mathbf{0}_k$ be the all-one and all-zero tensors with
the same shape as $\theta_T^{(k)}$. The corresponding mask block is
\begin{equation}
\mathcal M_{\mathcal E}^{(k)}
=
\begin{cases}
\mathbf{1}_k, & k\in\mathcal K_{\mathcal E},\\
\mathbf{0}_k, & k\notin\mathcal K_{\mathcal E}.
\end{cases}
\label{eq:supp_injection_mask}
\end{equation}

The alignment operator first maps the source model into the target parameter
space:
\begin{equation}
\widetilde{\Theta}_S
\triangleq
\mathcal A_{\pi}(\Theta_S).
\label{eq:supp_aligned_source}
\end{equation}
The projected direction and the final update are
\begin{align}
\Delta\Theta_{\mathcal E,\pi}
&=
\mathcal M_{\mathcal E}
\odot
\bigl(\widetilde{\Theta}_S-\Theta_T\bigr),
\label{eq:supp_masked_direction}\\
\theta_F^{(k)}
&=
\begin{cases}
(1-\alpha)\theta_T^{(k)}
+\alpha\widetilde{\theta}_S^{(k)},
& k\in\mathcal K_{\mathcal E},\\
\theta_T^{(k)}, & k\notin\mathcal K_{\mathcal E}.
\end{cases}
\label{eq:supp_parameter_update}
\end{align}
Thus, active tensors are interpolated with their aligned source counterparts,
whereas inactive target tensors are preserved exactly.

\subsection{Configuration Names Used in the Result Tables}

The result files use short internal names. In the tables below,
\texttt{all-linear} denotes $(C=\mathrm{All},P=\texttt{linear})$ with Prefix
mapping, and \texttt{all-merge} denotes
$(C=\mathrm{All},P=\texttt{full})$ with Prefix mapping.
\texttt{llm-linear-front}, \texttt{llm-linear-proportional},
\texttt{llm-linear-tail}, and \texttt{llm-linear-two-ends} correspond to
Prefix, Depth-proportional, Suffix, and Dual-end LLM mappings, respectively.

\section{Experimental and Evaluation Details}
\label{sec:supp_evaluation_details}

\subsection{Models and Configurations}

We use Qwen3-VL-Instruct models at four scales: 2B, 4B, 8B, and 32B.
The four evaluated source--target pairs are
4B$\rightarrow$2B, 8B$\rightarrow$2B,
32B$\rightarrow$4B, and 32B$\rightarrow$8B.
Table~\ref{tab:supp_model_pairs} gives the exact model identifiers used in
each direction.

\begin{table*}[!tbp]
\centering
\normalsize
\setlength{\tabcolsep}{12pt}
\begin{tabular}{@{}lll@{}}
\toprule
Direction & Source model ID & Target model ID \\
\midrule
4B$\rightarrow$2B
& \texttt{Qwen3-VL-4B-Instruct}
& \texttt{Qwen3-VL-2B-Instruct} \\
8B$\rightarrow$2B
& \texttt{Qwen3-VL-8B-Instruct}
& \texttt{Qwen3-VL-2B-Instruct} \\
32B$\rightarrow$4B
& \texttt{Qwen3-VL-32B-Instruct}
& \texttt{Qwen3-VL-4B-Instruct} \\
32B$\rightarrow$8B
& \texttt{Qwen3-VL-32B-Instruct}
& \texttt{Qwen3-VL-8B-Instruct} \\
\bottomrule
\end{tabular}
\caption{Exact model identifiers for the four evaluated source--target
directions.}
\label{tab:supp_model_pairs}
\end{table*}

Unless otherwise specified, CDPI uses $C=\mathrm{All}$,
$P=\texttt{linear}$, and Prefix LLM layer mapping. The main injection-ratio
grid is
\[
\alpha\in\{0.01,0.02,0.04,0.06,0.08\}.
\]
The additional ratios $0.16$ and $0.32$ are used only as stress tests for the
8B$\rightarrow$2B pair.

\subsection{Benchmark Groups}

The twelve benchmarks are grouped as follows:
\begin{itemize}
    \item \textbf{High-level reasoning:} MMMU-Pro, MathVista, MATH-Vision,
    VisuLogic, and VisualPuzzles.
    \item \textbf{General reasoning:} MMMU and MMVU.
    \item \textbf{Perception:} MME, MMStar, BLINK, OCRBench, and ChartQA.
\end{itemize}

We group benchmarks according to their dominant evaluation demands rather
than treating the three categories as mutually exclusive. Here,
perception-oriented benchmarks refer to tasks that place relatively greater
emphasis on direct visual understanding than the general- and high-level
reasoning groups, although some may also involve reasoning.
Every capability average is an unweighted arithmetic mean over the benchmarks
in the corresponding group. We use these full benchmark names consistently
throughout all result tables.

\subsection{Evaluation Settings}

All models are evaluated with the \texttt{qwen3\_vl} backend in
\texttt{lmms-eval}. Unless otherwise specified, we use the
benchmark-specific generation and scoring settings. MathVista and MATH-Vision
use local variants with a maximum generation length of 2048 tokens. Evaluation
uses rule-based answer extraction, exact matching, or the corresponding
standard evaluator supplied by \texttt{lmms-eval}; no GPT-based judging is
used. For MMVU, at most 32 frames are sampled uniformly from each video. The
experiments are conducted on a server equipped with eight NVIDIA A100 GPUs.

\section{Result Selection and Aggregation Protocol}
\label{sec:supp_reporting_protocol}

For benchmark $b$, let $S_{\alpha}(b)$ be the score obtained at injection
ratio $\alpha$, and let $S_T(b)$ be the target-model score. The fixed-ratio
gain is
\begin{equation}
\operatorname{Gain}_{\alpha}(b)
=
S_{\alpha}(b)-S_T(b).
\label{eq:supp_fixed_gain}
\end{equation}

The main benchmark-level tables use a benchmark-wise best-of-sweep summary:
\begin{equation}
\alpha_b^\star
\in
\arg\max_{\alpha\in\{0.01,0.02,0.04,0.06,0.08\}}
S_{\alpha}(b).
\label{eq:supp_best_alpha}
\end{equation}
If several ratios yield the same score, the smallest such ratio is displayed;
this convention changes only the displayed ratio, not the best score. The
best-of-sweep capability score for a benchmark group $\mathcal B$ is
\begin{equation}
\overline{S}_{\mathrm{best}}(\mathcal B)
=
\frac{1}{|\mathcal B|}
\sum_{b\in\mathcal B}
S_{\alpha_b^\star}(b).
\label{eq:supp_best_average}
\end{equation}
Consequently, a best-of-sweep capability average summarizes
benchmark-specific optima and does not correspond to one fused checkpoint.
For comparison, the fixed-ratio tables retain one common $\alpha$ across all
benchmarks. Averages are computed before rounding; displayed values are rounded
to two decimal places.

\section{Complete Benchmark-Wise Best-of-Sweep Results}
\label{sec:supp_best_results}

Tables~\ref{tab:supp_best_4b2b}--\ref{tab:supp_best_32b8b} report the unfused
source and target scores, best CDPI score, gain, and selected ratio for every
benchmark and model pair. The Average rows first select the best ratio
independently for each benchmark and then average the selected scores.

\begin{table}[!htbp]
\centering
\normalsize
\setlength{\tabcolsep}{3.7pt}
\begin{tabular}{@{}lrrrrr@{}}
\toprule
Benchmark & Source & Target & Best CDPI & Gain & $\alpha^\star$ \\
\midrule
MMMU-Pro & 31.79 & 23.76 & 25.32 & +1.56 & 0.04 \\
MathVista & 53.30 & 48.80 & 48.30 & -0.50 & 0.01 \\
MATH-Vision & 19.41 & 7.89 & 9.54 & +1.65 & 0.06 \\
VisuLogic & 26.80 & 16.70 & 26.00 & +9.30 & 0.08 \\
VisualPuzzles & 29.79 & 27.65 & 29.54 & +1.89 & 0.01 \\
\midrule
MMMU & 50.67 & 42.67 & 43.22 & +0.55 & 0.02 \\
MMVU & 52.50 & 50.20 & 51.30 & +1.10 & 0.01 \\
\midrule
MME & 83.86 & 71.80 & 72.20 & +0.40 & 0.02 \\
MMStar & 61.02 & 54.05 & 53.80 & -0.25 & 0.01 \\
BLINK & 63.92 & 43.50 & 43.13 & -0.37 & 0.01 \\
OCRBench & 83.10 & 80.90 & 80.60 & -0.30 & 0.01 \\
ChartQA & 83.72 & 79.68 & 79.80 & +0.12 & 0.01 \\
\midrule
\textbf{High-level avg.} & \textbf{32.22} & \textbf{24.96} & \textbf{27.74} & \textbf{+2.78} & -- \\
\textbf{General avg.} & \textbf{51.59} & \textbf{46.44} & \textbf{47.26} & \textbf{+0.82} & -- \\
\textbf{Perception avg.} & \textbf{75.12} & \textbf{65.99} & \textbf{65.91} & \textbf{-0.08} & -- \\
\bottomrule
\end{tabular}
\caption{Benchmark-wise best-of-sweep results for 4B$\rightarrow$2B. Source and
Target are the unfused model baselines; Gain is Best CDPI minus Target.}
\label{tab:supp_best_4b2b}
\end{table}

\begin{table}[!htbp]
\centering
\normalsize
\setlength{\tabcolsep}{3.7pt}
\begin{tabular}{@{}lrrrrr@{}}
\toprule
Benchmark & Source & Target & Best CDPI & Gain & $\alpha^\star$ \\
\midrule
MMMU-Pro & 39.31 & 23.76 & 24.57 & +0.81 & 0.02 \\
MathVista & 59.60 & 48.80 & 48.30 & -0.50 & 0.01 \\
MATH-Vision & 26.32 & 7.89 & 11.18 & +3.29 & 0.02 \\
VisuLogic & 27.00 & 16.70 & 26.20 & +9.50 & 0.04 \\
VisualPuzzles & 31.76 & 27.65 & 30.05 & +2.40 & 0.02 \\
\midrule
MMMU & 52.89 & 42.67 & 44.11 & +1.44 & 0.06 \\
MMVU & 57.30 & 50.20 & 50.60 & +0.40 & 0.02 \\
\midrule
MME & 84.62 & 71.80 & 72.08 & +0.28 & 0.01 \\
MMStar & 63.87 & 54.05 & 53.87 & -0.18 & 0.04 \\
BLINK & 65.05 & 43.50 & 43.97 & +0.47 & 0.01 \\
OCRBench & 85.00 & 80.90 & 80.80 & -0.10 & 0.01 \\
ChartQA & 84.96 & 79.68 & 79.88 & +0.20 & 0.01 \\
\midrule
\textbf{High-level avg.} & \textbf{36.80} & \textbf{24.96} & \textbf{28.06} & \textbf{+3.10} & -- \\
\textbf{General avg.} & \textbf{55.09} & \textbf{46.44} & \textbf{47.36} & \textbf{+0.92} & -- \\
\textbf{Perception avg.} & \textbf{76.70} & \textbf{65.99} & \textbf{66.12} & \textbf{+0.13} & -- \\
\bottomrule
\end{tabular}
\caption{Benchmark-wise best-of-sweep results for 8B$\rightarrow$2B. Source and
Target are the unfused model baselines; Gain is Best CDPI minus Target.}
\label{tab:supp_best_8b2b}
\end{table}

\begin{table}[!htbp]
\centering
\normalsize
\setlength{\tabcolsep}{3.7pt}
\begin{tabular}{@{}lrrrrr@{}}
\toprule
Benchmark & Source & Target & Best CDPI & Gain & $\alpha^\star$ \\
\midrule
MMMU-Pro & 45.61 & 31.79 & 33.74 & +1.95 & 0.01 \\
MathVista & 61.50 & 53.30 & 55.30 & +2.00 & 0.02 \\
MATH-Vision & 33.22 & 19.41 & 23.81 & +4.40 & 0.02 \\
VisuLogic & 24.80 & 26.80 & 27.10 & +0.30 & 0.04 \\
VisualPuzzles & 33.39 & 29.79 & 31.74 & +1.95 & 0.02 \\
\midrule
MMMU & 60.67 & 50.67 & 51.33 & +0.66 & 0.04 \\
MMVU & 64.50 & 52.50 & 53.90 & +1.40 & 0.02 \\
\midrule
MME & 88.31 & 83.86 & 84.53 & +0.67 & 0.01 \\
MMStar & 70.56 & 61.02 & 61.76 & +0.74 & 0.02 \\
BLINK & 67.53 & 63.92 & 64.26 & +0.34 & 0.01 \\
OCRBench & 86.20 & 83.10 & 82.60 & -0.50 & 0.01 \\
ChartQA & 83.44 & 83.72 & 83.68 & -0.04 & 0.01 \\
\midrule
\textbf{High-level avg.} & \textbf{39.70} & \textbf{32.22} & \textbf{34.34} & \textbf{+2.12} & -- \\
\textbf{General avg.} & \textbf{62.59} & \textbf{51.59} & \textbf{52.62} & \textbf{+1.03} & -- \\
\textbf{Perception avg.} & \textbf{79.21} & \textbf{75.12} & \textbf{75.37} & \textbf{+0.24} & -- \\
\bottomrule
\end{tabular}
\caption{Benchmark-wise best-of-sweep results for 32B$\rightarrow$4B. Source and
Target are the unfused model baselines; Gain is Best CDPI minus Target.}
\label{tab:supp_best_32b4b}
\end{table}

\begin{table}[!htbp]
\centering
\normalsize
\setlength{\tabcolsep}{3.7pt}
\begin{tabular}{@{}lrrrrr@{}}
\toprule
Benchmark & Source & Target & Best CDPI & Gain & $\alpha^\star$ \\
\midrule
MMMU-Pro & 45.61 & 39.31 & 39.94 & +0.63 & 0.02 \\
MathVista & 61.50 & 59.60 & 60.40 & +0.80 & 0.06 \\
MATH-Vision & 33.22 & 26.32 & 27.96 & +1.64 & 0.01 \\
VisuLogic & 24.80 & 27.00 & 26.80 & -0.20 & 0.01 \\
VisualPuzzles & 33.39 & 31.76 & 34.08 & +2.32 & 0.08 \\
\midrule
MMMU & 60.67 & 52.89 & 53.89 & +1.00 & 0.02 \\
MMVU & 64.50 & 57.30 & 59.30 & +2.00 & 0.01 \\
\midrule
MME & 88.31 & 84.62 & 84.89 & +0.27 & 0.01 \\
MMStar & 70.56 & 63.87 & 64.16 & +0.29 & 0.01 \\
BLINK & 67.53 & 65.05 & 65.22 & +0.17 & 0.01 \\
OCRBench & 86.20 & 85.00 & 85.30 & +0.30 & 0.02 \\
ChartQA & 83.44 & 84.96 & 84.84 & -0.12 & 0.01 \\
\midrule
\textbf{High-level avg.} & \textbf{39.70} & \textbf{36.80} & \textbf{37.84} & \textbf{+1.04} & -- \\
\textbf{General avg.} & \textbf{62.59} & \textbf{55.09} & \textbf{56.59} & \textbf{+1.50} & -- \\
\textbf{Perception avg.} & \textbf{79.21} & \textbf{76.70} & \textbf{76.88} & \textbf{+0.18} & -- \\
\bottomrule
\end{tabular}
\caption{Benchmark-wise best-of-sweep results for 32B$\rightarrow$8B. Source and
Target are the unfused model baselines; Gain is Best CDPI minus Target.}
\label{tab:supp_best_32b8b}
\end{table}

\FloatBarrier

\section{Complete Fixed-Ratio Results}
\label{sec:supp_fixed_results}

This section reports all twelve benchmark scores at every ratio in the main
search grid. Unlike the preceding best-of-sweep tables, each row here
is a benchmark and each ratio column corresponds to one fused checkpoint.
The Source and Target columns report the two unfused model baselines. In each
capability-average row, every ratio entry is the unweighted
average within that ratio column, while the Best entry is the unweighted
average of the benchmark-wise values in the Best column. Thus, the Best
average first selects the best ratio independently for each benchmark and
then averages those selected scores, matching the aggregation used in
Section~\ref{sec:supp_best_results}. Boldface in the ratio columns marks every
value attaining the row maximum; ties are all bolded.

\begin{table*}[!tbp]
\centering
\normalsize
\setlength{\tabcolsep}{5.5pt}
\begin{tabular}{@{}lrrrrrrrr@{}}
\toprule
Benchmark & Source & Target & 0.01 & 0.02 & 0.04 & 0.06 & 0.08 & Best \\
\midrule
MMMU-Pro & 31.79 & 23.76 & 24.86 & 25.15 & \textbf{25.32} & 24.22 & 20.98 & \textbf{25.32} \\
MathVista & 53.30 & 48.80 & \textbf{48.30} & 47.70 & 46.10 & 45.00 & 43.00 & \textbf{48.30} \\
MATH-Vision & 19.41 & 7.89 & 8.22 & 8.22 & 7.57 & \textbf{9.54} & 7.89 & \textbf{9.54} \\
VisuLogic & 26.80 & 16.70 & 16.00 & 17.40 & 20.50 & 21.80 & \textbf{26.00} & \textbf{26.00} \\
VisualPuzzles & 29.79 & 27.65 & \textbf{29.54} & 27.83 & 26.71 & 26.63 & 27.74 & \textbf{29.54} \\
\midrule
MMMU & 50.67 & 42.67 & 42.89 & \textbf{43.22} & 42.78 & 43.11 & 42.33 & \textbf{43.22} \\
MMVU & 52.50 & 50.20 & \textbf{51.30} & 48.60 & 48.00 & 46.80 & 44.50 & \textbf{51.30} \\
\midrule
High-level avg. & 32.22 & 24.96 & 25.38 & 25.26 & 25.24 & \textbf{25.44} & 25.12 & \textbf{27.74} \\
General avg. & 51.59 & 46.44 & \textbf{47.09} & 45.91 & 45.39 & 44.95 & 43.41 & \textbf{47.26} \\
\bottomrule
\end{tabular}
\caption{Fixed-ratio reasoning results for 4B$\rightarrow$2B. Best is the maximum
CDPI score per benchmark; average-row Best is the mean of these
benchmark-wise maxima.}
\label{tab:supp_fixed_reason_4b2b}
\end{table*}

\begin{table*}[!tbp]
\centering
\normalsize
\setlength{\tabcolsep}{5.5pt}
\begin{tabular}{@{}lrrrrrrrr@{}}
\toprule
Benchmark & Source & Target & 0.01 & 0.02 & 0.04 & 0.06 & 0.08 & Best \\
\midrule
MME & 83.86 & 71.80 & 72.10 & \textbf{72.20} & 72.11 & 71.18 & 68.25 & \textbf{72.20} \\
MMStar & 61.02 & 54.05 & \textbf{53.80} & 53.13 & 50.83 & 50.52 & 49.05 & \textbf{53.80} \\
BLINK & 63.92 & 43.50 & \textbf{43.13} & 42.80 & 42.38 & 42.09 & 39.08 & \textbf{43.13} \\
OCRBench & 83.10 & 80.90 & \textbf{80.60} & 80.30 & 79.20 & 78.40 & 76.80 & \textbf{80.60} \\
ChartQA & 83.72 & 79.68 & \textbf{79.80} & 79.48 & 78.04 & 76.52 & 72.40 & \textbf{79.80} \\
\midrule
Perception avg. & 75.12 & 65.99 & \textbf{65.89} & 65.58 & 64.51 & 63.74 & 61.12 & \textbf{65.91} \\
\bottomrule
\end{tabular}
\caption{Fixed-ratio perception results for 4B$\rightarrow$2B. Best is the maximum
CDPI score per benchmark; average-row Best is the mean of these
benchmark-wise maxima.}
\label{tab:supp_fixed_perception_4b2b}
\end{table*}

\begin{table*}[!tbp]
\centering
\normalsize
\setlength{\tabcolsep}{5.5pt}
\begin{tabular}{@{}lrrrrrrrr@{}}
\toprule
Benchmark & Source & Target & 0.01 & 0.02 & 0.04 & 0.06 & 0.08 & Best \\
\midrule
MMMU-Pro & 39.31 & 23.76 & 23.99 & \textbf{24.57} & 22.95 & 20.93 & 15.15 & \textbf{24.57} \\
MathVista & 59.60 & 48.80 & \textbf{48.30} & 46.80 & 43.10 & 40.20 & 36.80 & \textbf{48.30} \\
MATH-Vision & 26.32 & 7.89 & 8.22 & \textbf{11.18} & 9.87 & 5.92 & 5.92 & \textbf{11.18} \\
VisuLogic & 27.00 & 16.70 & 18.00 & 20.70 & \textbf{26.20} & 22.70 & 20.90 & \textbf{26.20} \\
VisualPuzzles & 31.76 & 27.65 & 27.65 & \textbf{30.05} & 29.11 & 29.37 & 29.45 & \textbf{30.05} \\
\midrule
MMMU & 52.89 & 42.67 & 43.00 & 42.67 & 42.44 & \textbf{44.11} & 41.67 & \textbf{44.11} \\
MMVU & 57.30 & 50.20 & 48.90 & \textbf{50.60} & 49.10 & \textbf{50.60} & 47.20 & \textbf{50.60} \\
\midrule
High-level avg. & 36.80 & 24.96 & 25.23 & \textbf{26.66} & 26.25 & 23.82 & 21.64 & \textbf{28.06} \\
General avg. & 55.09 & 46.44 & 45.95 & 46.64 & 45.77 & \textbf{47.36} & 44.44 & \textbf{47.36} \\
\bottomrule
\end{tabular}
\caption{Fixed-ratio reasoning results for 8B$\rightarrow$2B. Best is the maximum
CDPI score per benchmark; average-row Best is the mean of these
benchmark-wise maxima.}
\label{tab:supp_fixed_reason_8b2b}
\end{table*}

\begin{table*}[!tbp]
\centering
\normalsize
\setlength{\tabcolsep}{5.5pt}
\begin{tabular}{@{}lrrrrrrrr@{}}
\toprule
Benchmark & Source & Target & 0.01 & 0.02 & 0.04 & 0.06 & 0.08 & Best \\
\midrule
MME & 84.62 & 71.80 & \textbf{72.08} & 71.94 & 70.14 & 67.13 & 61.37 & \textbf{72.08} \\
MMStar & 63.87 & 54.05 & 53.56 & 53.85 & \textbf{53.87} & 51.33 & 48.55 & \textbf{53.87} \\
BLINK & 65.05 & 43.50 & \textbf{43.97} & 43.41 & 43.66 & 43.25 & 43.62 & \textbf{43.97} \\
OCRBench & 85.00 & 80.90 & \textbf{80.80} & 79.90 & 77.20 & 74.10 & 66.60 & \textbf{80.80} \\
ChartQA & 84.96 & 79.68 & \textbf{79.88} & 79.40 & 78.28 & 75.04 & 67.96 & \textbf{79.88} \\
\midrule
Perception avg. & 76.70 & 65.99 & \textbf{66.06} & 65.70 & 64.63 & 62.17 & 57.62 & \textbf{66.12} \\
\bottomrule
\end{tabular}
\caption{Fixed-ratio perception results for 8B$\rightarrow$2B. Best is the maximum
CDPI score per benchmark; average-row Best is the mean of these
benchmark-wise maxima.}
\label{tab:supp_fixed_perception_8b2b}
\end{table*}

\begin{table*}[!tbp]
\centering
\normalsize
\setlength{\tabcolsep}{5.5pt}
\begin{tabular}{@{}lrrrrrrrr@{}}
\toprule
Benchmark & Source & Target & 0.01 & 0.02 & 0.04 & 0.06 & 0.08 & Best \\
\midrule
MMMU-Pro & 45.61 & 31.79 & \textbf{33.74} & 33.24 & 32.90 & 31.62 & 28.96 & \textbf{33.74} \\
MathVista & 61.50 & 53.30 & 54.60 & \textbf{55.30} & 54.10 & 51.10 & 50.10 & \textbf{55.30} \\
MATH-Vision & 33.22 & 19.41 & 19.06 & \textbf{23.81} & 22.64 & 21.05 & 20.07 & \textbf{23.81} \\
VisuLogic & 24.80 & 26.80 & 26.30 & 26.40 & \textbf{27.10} & 26.20 & 26.70 & \textbf{27.10} \\
VisualPuzzles & 33.39 & 29.79 & 29.02 & \textbf{31.74} & 30.48 & 30.05 & 31.68 & \textbf{31.74} \\
\midrule
MMMU & 60.67 & 50.67 & 50.44 & 51.00 & \textbf{51.33} & 50.33 & 49.67 & \textbf{51.33} \\
MMVU & 64.50 & 52.50 & 53.10 & \textbf{53.90} & 53.10 & 52.00 & 49.80 & \textbf{53.90} \\
\midrule
High-level avg. & 39.70 & 32.22 & 32.54 & \textbf{34.10} & 33.44 & 32.00 & 31.50 & \textbf{34.34} \\
General avg. & 62.59 & 51.59 & 51.77 & \textbf{52.45} & 52.22 & 51.17 & 49.74 & \textbf{52.62} \\
\bottomrule
\end{tabular}
\caption{Fixed-ratio reasoning results for 32B$\rightarrow$4B. Best is the maximum
CDPI score per benchmark; average-row Best is the mean of these
benchmark-wise maxima.}
\label{tab:supp_fixed_reason_32b4b}
\end{table*}

\begin{table*}[!tbp]
\centering
\normalsize
\setlength{\tabcolsep}{5.5pt}
\begin{tabular}{@{}lrrrrrrrr@{}}
\toprule
Benchmark & Source & Target & 0.01 & 0.02 & 0.04 & 0.06 & 0.08 & Best \\
\midrule
MME & 88.31 & 83.86 & \textbf{84.53} & 83.88 & 81.84 & 81.24 & 78.43 & \textbf{84.53} \\
MMStar & 70.56 & 61.02 & 61.13 & \textbf{61.76} & 60.96 & 59.76 & 58.77 & \textbf{61.76} \\
BLINK & 67.53 & 63.92 & \textbf{64.26} & 63.78 & 63.31 & 61.57 & 57.99 & \textbf{64.26} \\
OCRBench & 86.20 & 83.10 & \textbf{82.60} & 81.90 & 80.50 & 80.10 & 74.90 & \textbf{82.60} \\
ChartQA & 83.44 & 83.72 & \textbf{83.68} & 83.48 & 81.64 & 80.00 & 74.60 & \textbf{83.68} \\
\midrule
Perception avg. & 79.21 & 75.12 & \textbf{75.24} & 74.96 & 73.65 & 72.53 & 68.94 & \textbf{75.37} \\
\bottomrule
\end{tabular}
\caption{Fixed-ratio perception results for 32B$\rightarrow$4B. Best is the maximum
CDPI score per benchmark; average-row Best is the mean of these
benchmark-wise maxima.}
\label{tab:supp_fixed_perception_32b4b}
\end{table*}

\begin{table*}[!tbp]
\centering
\normalsize
\setlength{\tabcolsep}{5.5pt}
\begin{tabular}{@{}lrrrrrrrr@{}}
\toprule
Benchmark & Source & Target & 0.01 & 0.02 & 0.04 & 0.06 & 0.08 & Best \\
\midrule
MMMU-Pro & 45.61 & 39.31 & 39.60 & \textbf{39.94} & 39.36 & 37.98 & 36.76 & \textbf{39.94} \\
MathVista & 61.50 & 59.60 & 59.80 & 59.70 & 60.20 & \textbf{60.40} & 59.70 & \textbf{60.40} \\
MATH-Vision & 33.22 & 26.32 & \textbf{27.96} & 26.97 & \textbf{27.96} & 26.32 & 24.01 & \textbf{27.96} \\
VisuLogic & 24.80 & 27.00 & \textbf{26.80} & 26.20 & 25.90 & 26.00 & 25.80 & \textbf{26.80} \\
VisualPuzzles & 33.39 & 31.76 & 32.62 & 32.53 & 32.88 & 33.99 & \textbf{34.08} & \textbf{34.08} \\
\midrule
MMMU & 60.67 & 52.89 & 53.33 & \textbf{53.89} & 52.33 & 52.11 & 51.89 & \textbf{53.89} \\
MMVU & 64.50 & 57.30 & \textbf{59.30} & 59.10 & 57.70 & 57.30 & 56.30 & \textbf{59.30} \\
\midrule
High-level avg. & 39.70 & 36.80 & \textbf{37.36} & 37.07 & 37.26 & 36.94 & 36.07 & \textbf{37.84} \\
General avg. & 62.59 & 55.09 & 56.31 & \textbf{56.50} & 55.02 & 54.70 & 54.09 & \textbf{56.59} \\
\bottomrule
\end{tabular}
\caption{Fixed-ratio reasoning results for 32B$\rightarrow$8B. Best is the maximum
CDPI score per benchmark; average-row Best is the mean of these
benchmark-wise maxima.}
\label{tab:supp_fixed_reason_32b8b}
\end{table*}

\begin{table*}[!tbp]
\centering
\normalsize
\setlength{\tabcolsep}{5.5pt}
\begin{tabular}{@{}lrrrrrrrr@{}}
\toprule
Benchmark & Source & Target & 0.01 & 0.02 & 0.04 & 0.06 & 0.08 & Best \\
\midrule
MME & 88.31 & 84.62 & \textbf{84.89} & 84.08 & 84.24 & 83.30 & 82.46 & \textbf{84.89} \\
MMStar & 70.56 & 63.87 & \textbf{64.16} & 63.32 & 63.35 & 63.46 & 61.97 & \textbf{64.16} \\
BLINK & 67.53 & 65.05 & \textbf{65.22} & 64.61 & 63.88 & 62.26 & 60.66 & \textbf{65.22} \\
OCRBench & 86.20 & 85.00 & 84.80 & \textbf{85.30} & 84.70 & 84.70 & 84.80 & \textbf{85.30} \\
ChartQA & 83.44 & 84.96 & \textbf{84.84} & 84.52 & 83.96 & 83.08 & 81.20 & \textbf{84.84} \\
\midrule
Perception avg. & 79.21 & 76.70 & \textbf{76.78} & 76.37 & 76.03 & 75.36 & 74.22 & \textbf{76.88} \\
\bottomrule
\end{tabular}
\caption{Fixed-ratio perception results for 32B$\rightarrow$8B. Best is the maximum
CDPI score per benchmark; average-row Best is the mean of these
benchmark-wise maxima.}
\label{tab:supp_fixed_perception_32b8b}
\end{table*}

\FloatBarrier

\subsection{Stress-Test Ratios for 8B$\rightarrow$2B}

The ratios $\alpha=0.16$ and $\alpha=0.32$ lie outside the main search grid
and are included only to examine degradation under larger parameter movement.
At $\alpha=0.16$, the MMVU output is affected by a scoring artifact; its raw
score is shown for transparency but excluded from the General average.

\begin{table}[!htbp]
\centering
\normalsize
\setlength{\tabcolsep}{4pt}
\begin{tabular}{@{}lrrrr@{}}
\toprule
Benchmark & Source & Target & 0.16 & 0.32 \\
\midrule
MMMU-Pro & 39.31 & 23.76 & 0.00 & 0.00 \\
MathVista & 59.60 & 48.80 & 17.30 & 18.30 \\
MATH-Vision & 26.32 & 7.89 & 0.33 & 0.00 \\
VisuLogic & 27.00 & 16.70 & 0.30 & 0.00 \\
VisualPuzzles & 31.76 & 27.65 & 22.69 & 25.34 \\
\midrule
MMMU & 52.89 & 42.67 & 27.00 & 22.33 \\
MMVU & 57.30 & 50.20 & 58.40$^{\dagger}$ & 0.60 \\
\midrule
MME & 84.62 & 71.80 & 24.71 & 0.00 \\
MMStar & 63.87 & 54.05 & 26.98 & 0.00 \\
BLINK & 65.05 & 43.50 & 37.26 & 0.00 \\
OCRBench & 85.00 & 80.90 & 2.50 & 0.00 \\
ChartQA & 84.96 & 79.68 & 7.24 & 0.00 \\
\midrule
\textbf{High-level avg.} & \textbf{36.80} & \textbf{24.96} & \textbf{8.12} & \textbf{8.73} \\
\textbf{General avg.} & \textbf{55.09} & \textbf{46.44} & -- & \textbf{11.46} \\
\textbf{Perception avg.} & \textbf{76.70} & \textbf{65.99} & \textbf{19.74} & \textbf{0.00} \\
\bottomrule
\end{tabular}
\caption{Complete 8B$\rightarrow$2B stress-test results, including the unfused
Source and Target baselines. The MMVU value marked by $\dagger$ is affected by
a scoring artifact and is excluded from the General average.}
\label{tab:supp_stress_complete}
\end{table}

\FloatBarrier

\section{Complete Ablation Summaries}
\label{sec:supp_ablation_results}

The following tables report capability gains derived from the complete
benchmark-wise best-of-sweep matrices. The same benchmark grouping and
aggregation rules from Section~\ref{sec:supp_reporting_protocol} are used.

\subsection{Parameter Scope}

\begin{table}[!htbp]
\centering
\normalsize
\setlength{\tabcolsep}{4pt}
\begin{tabular}{@{}llrrr@{}}
\toprule
Pair & Setting & High-level & General & Perception \\
\midrule
4B$\rightarrow$2B & Linear & +2.78 & +0.82 & -0.08 \\
 & Full & +0.75 & -1.19 & -0.44 \\
\midrule
8B$\rightarrow$2B & Linear & +3.10 & +0.92 & +0.13 \\
 & Full & +1.61 & +0.20 & +0.22 \\
\midrule
32B$\rightarrow$4B & Linear & +2.12 & +1.03 & +0.24 \\
 & Full & +1.81 & +0.09 & +0.06 \\
\midrule
32B$\rightarrow$8B & Linear & +1.04 & +1.50 & +0.18 \\
 & Full & +1.02 & +0.72 & +0.05 \\
\bottomrule
\end{tabular}
\caption{Benchmark-wise best-of-sweep capability gains for the two parameter scopes. Linear selects linear-layer weights, while Full includes all alignable parameters in the selected components.}
\label{tab:supp_scope_gains}
\end{table}

\subsection{Injected Component}

\begin{table}[!htbp]
\centering
\normalsize
\setlength{\tabcolsep}{4pt}
\begin{tabular}{@{}llrrr@{}}
\toprule
Pair & Setting & High-level & General & Perception \\
\midrule
4B$\rightarrow$2B & All & +2.78 & +0.82 & -0.08 \\
 & ViT & +0.92 & +0.41 & +0.44 \\
 & Proj & +0.46 & +0.55 & +0.29 \\
 & LLM & +2.78 & -0.02 & -0.14 \\
\midrule
8B$\rightarrow$2B & All & +3.10 & +0.92 & +0.13 \\
 & ViT & +1.10 & +1.11 & +0.54 \\
 & Proj & +0.63 & +1.01 & +0.14 \\
 & LLM & +2.96 & +1.16 & -0.09 \\
\midrule
32B$\rightarrow$4B & All & +2.12 & +1.03 & +0.24 \\
 & ViT & +0.83 & +0.42 & +0.26 \\
 & Proj & +1.01 & +1.09 & +0.13 \\
 & LLM & +2.62 & +0.55 & -0.01 \\
\midrule
32B$\rightarrow$8B & All & +1.04 & +1.50 & +0.18 \\
 & ViT & +0.57 & +1.04 & +0.28 \\
 & Proj & +0.12 & +1.54 & +0.27 \\
 & LLM & +1.69 & +0.67 & -0.08 \\
\bottomrule
\end{tabular}
\caption{Benchmark-wise best-of-sweep capability gains under component-wise linear injection. Prefix mapping is used for LLM and All.}
\label{tab:supp_component_gains}
\end{table}

\subsection{LLM Layer Mapping}

\begin{table}[!htbp]
\centering
\normalsize
\setlength{\tabcolsep}{4pt}
\begin{tabular}{@{}llrrr@{}}
\toprule
Pair & Setting & High-level & General & Perception \\
\midrule
4B$\rightarrow$2B & Prefix & +2.78 & -0.02 & -0.14 \\
 & Depth-prop. & +2.33 & +0.33 & +1.05 \\
 & Suffix & +3.08 & +2.25 & +0.67 \\
 & Dual-end & +3.08 & +0.61 & -0.03 \\
\midrule
8B$\rightarrow$2B & Prefix & +2.96 & +1.16 & -0.09 \\
 & Depth-prop. & +2.60 & +0.49 & +0.63 \\
 & Suffix & +3.02 & +0.63 & +0.77 \\
 & Dual-end & +2.59 & +1.52 & +0.08 \\
\midrule
32B$\rightarrow$4B & Prefix & +2.62 & +0.55 & -0.01 \\
 & Depth-prop. & +1.27 & +0.70 & +0.35 \\
 & Suffix & +1.26 & +0.61 & +0.05 \\
 & Dual-end & +1.50 & +0.75 & +0.10 \\
\midrule
32B$\rightarrow$8B & Prefix & +1.69 & +0.67 & -0.08 \\
 & Depth-prop. & +1.33 & +1.41 & +0.40 \\
 & Suffix & +1.55 & +0.80 & +0.06 \\
 & Dual-end & +1.73 & +1.09 & +0.05 \\
\bottomrule
\end{tabular}
\caption{Benchmark-wise best-of-sweep capability gains for the four LLM layer mappings under linear LLM-only injection.}
\label{tab:supp_mapping_gains}
\end{table}

\FloatBarrier

\section{Random Perturbation Control}
\label{sec:supp_random_perturbation}

We use a random perturbation control to test whether the improvements produced
by CDPI can be reproduced by a generic parameter perturbation containing no
information from the source model. Consistent with the injection-ratio
analysis, we use the 8B$\rightarrow$2B setting and perturb the same
$C=\mathrm{All}$, $P=\texttt{linear}$ parameter subset.

For each selected target weight tensor $W$ and random seed $s$, we sample an
independent multiplicative mask with the same shape as $W$:
\begin{equation}
R_{ij}^{(s)}
\overset{\mathrm{i.i.d.}}{\sim}
\mathcal U(1-\delta,1+\delta),
\qquad
\delta=0.08.
\label{eq:supp_random_mask}
\end{equation}
The perturbed tensor is
\begin{equation}
W_{\mathrm{rand}}^{(s)}
=
W\odot R^{(s)}.
\label{eq:supp_random_update}
\end{equation}
Thus, every selected weight is multiplied by a factor sampled from
$[0.92,1.08]$. Because $\mathbb E[R_{ij}^{(s)}]=1$, the perturbation is
zero-mean relative to the original target tensor:
\begin{equation}
\mathbb E\!\left[W_{\mathrm{rand}}^{(s)}-W\right]=0.
\label{eq:supp_random_zero_mean}
\end{equation}
To make the perturbation scale explicit, define
$\varepsilon_{ij}^{(s)}=R_{ij}^{(s)}-1$ and
$\Delta W^{(s)}=W_{\mathrm{rand}}^{(s)}-W
=W\odot\varepsilon^{(s)}$. Then
\begin{equation}
\operatorname{Var}\!\left(\varepsilon_{ij}^{(s)}\right)
=
\frac{\delta^2}{3}
=
\frac{0.08^2}{3}.
\label{eq:supp_random_variance}
\end{equation}
Using the independence and zero mean of the entries of
$\varepsilon^{(s)}$, the expected squared perturbation norm is
\begin{equation}
\mathbb E\!\left[
\left\lVert\Delta W^{(s)}\right\rVert_F^2
\right]
=
\frac{0.08^2}{3}
\left\lVert W\right\rVert_F^2.
\label{eq:supp_random_expected_norm}
\end{equation}
The corresponding root-mean-square relative perturbation strength is
\begin{equation}
\frac{
\sqrt{
\mathbb E\!\left[
\left\lVert\Delta W^{(s)}\right\rVert_F^2
\right]
}
}{
\left\lVert W\right\rVert_F
}
=
\frac{0.08}{\sqrt{3}}
\approx
0.0462
=
4.62\%.
\label{eq:supp_random_relative_strength}
\end{equation}
This places the control in a few-percent perturbation regime, making it a
useful practical scale comparison for CDPI at $\alpha=0.02$ and
$\alpha=0.04$. This is a scale comparison rather than exact norm matching,
because CDPI scales an aligned source--target direction instead of applying a
multiplicative perturbation to $W$.
Unlike CDPI, this control neither uses aligned source parameters nor defines a
source--target transfer direction.

For this perturbation experiment, we evaluate five random seeds:
$11$, $42$, $67$, $527$, and $1483$. Each seed produces one independently
perturbed target model. We report the per-seed scores together with their mean
and sample standard deviation, and compare them with CDPI at
$\alpha=0.02$ and $\alpha=0.04$. No training, fine-tuning, or gradient-based
optimization is performed.

In Tables~\ref{tab:supp_random_high_level}--%
\ref{tab:supp_random_perception_cont}, \textit{Rnd.} denotes the random
perturbation control, and $\mu\pm\sigma$ denotes its mean and sample standard
deviation over the five seeds.

\begin{table}[!htbp]
\centering
\normalsize
\setlength{\tabcolsep}{3pt}
\begin{tabular}{@{}lrrr@{}}
\toprule
Method & MMMU-Pro & MathVista & MATH-Vision \\
\midrule
Target & 23.76 & 48.80 & 7.89 \\
CDPI ($0.02$) & 24.57 & 46.80 & 11.18 \\
CDPI ($0.04$) & 22.95 & 43.10 & 9.87 \\
\midrule
Rnd. seed 11 & 18.27 & 47.50 & 9.21 \\
Rnd. seed 42 & 25.32 & 47.80 & 9.54 \\
Rnd. seed 67 & 22.31 & 46.60 & 9.87 \\
Rnd. seed 527 & 24.45 & 48.10 & 9.21 \\
Rnd. seed 1483 & 22.95 & 47.40 & 8.88 \\
\midrule
Rnd. $\mu\!\pm\!\sigma$ & $22.66\!\pm\!2.73$ & $47.48\!\pm\!0.56$ & $9.34\!\pm\!0.38$ \\
\bottomrule
\end{tabular}
\caption{Random-control high-level reasoning results (part 1 of 2) for 8B$\rightarrow$2B.}
\label{tab:supp_random_high_level}
\end{table}

\begin{table}[!htbp]
\centering
\normalsize
\setlength{\tabcolsep}{3pt}
\begin{tabular}{@{}lrrr@{}}
\toprule
Method & VisuLogic & VisualPuzzles & High-level \\
\midrule
Target & 16.70 & 27.65 & 24.96 \\
CDPI ($0.02$) & 20.70 & 30.05 & 26.66 \\
CDPI ($0.04$) & 26.20 & 29.11 & 26.25 \\
\midrule
Rnd. seed 11 & 8.90 & 27.65 & 22.31 \\
Rnd. seed 42 & 21.70 & 24.40 & 25.75 \\
Rnd. seed 67 & 18.20 & 26.46 & 24.69 \\
Rnd. seed 527 & 6.00 & 28.77 & 23.31 \\
Rnd. seed 1483 & 13.80 & 27.14 & 24.03 \\
\midrule
Rnd. $\mu\!\pm\!\sigma$ & $13.72\!\pm\!6.45$ & $26.88\!\pm\!1.62$ & $24.02\!\pm\!1.31$ \\
\bottomrule
\end{tabular}
\caption{Random-control high-level reasoning results (part 2 of 2). High-level averages all five benchmarks.}
\label{tab:supp_random_high_level_cont}
\end{table}

\begin{table}[!htbp]
\centering
\normalsize
\setlength{\tabcolsep}{3pt}
\begin{tabular}{@{}lrrr@{}}
\toprule
Method & MMMU & MMVU & General \\
\midrule
Target & 42.67 & 50.20 & 46.44 \\
CDPI ($0.02$) & 42.67 & 50.60 & 46.64 \\
CDPI ($0.04$) & 42.44 & 49.10 & 45.77 \\
\midrule
Rnd. seed 11 & 43.67 & 46.50 & 45.08 \\
Rnd. seed 42 & 42.56 & 48.90 & 45.73 \\
Rnd. seed 67 & 42.11 & 47.60 & 44.86 \\
Rnd. seed 527 & 41.11 & 49.70 & 45.41 \\
Rnd. seed 1483 & 42.56 & 47.20 & 44.88 \\
\midrule
Rnd. $\mu\!\pm\!\sigma$ & $42.40\!\pm\!0.92$ & $47.98\!\pm\!1.30$ & $45.19\!\pm\!0.37$ \\
\bottomrule
\end{tabular}
\caption{Random-control general-reasoning results for 8B$\rightarrow$2B.}
\label{tab:supp_random_general}
\end{table}

\begin{table}[!htbp]
\centering
\normalsize
\setlength{\tabcolsep}{3pt}
\begin{tabular}{@{}lrrr@{}}
\toprule
Method & MME & MMStar & BLINK \\
\midrule
Target & 71.80 & 54.05 & 43.50 \\
CDPI ($0.02$) & 71.94 & 53.85 & 43.41 \\
CDPI ($0.04$) & 70.14 & 53.87 & 43.66 \\
\midrule
Rnd. seed 11 & 70.14 & 53.30 & 42.81 \\
Rnd. seed 42 & 72.96 & 53.78 & 43.96 \\
Rnd. seed 67 & 71.96 & 51.98 & 43.33 \\
Rnd. seed 527 & 70.83 & 53.80 & 41.77 \\
Rnd. seed 1483 & 69.25 & 52.48 & 45.05 \\
\midrule
Rnd. $\mu\!\pm\!\sigma$ & $71.03\!\pm\!1.46$ & $53.07\!\pm\!0.81$ & $43.38\!\pm\!1.23$ \\
\bottomrule
\end{tabular}
\caption{Random-control perception results (part 1 of 2) for 8B$\rightarrow$2B.}
\label{tab:supp_random_perception}
\end{table}

\begin{table}[!htbp]
\centering
\normalsize
\setlength{\tabcolsep}{3pt}
\begin{tabular}{@{}lrrr@{}}
\toprule
Method & OCRBench & ChartQA & Perception \\
\midrule
Target & 80.90 & 79.68 & 65.99 \\
CDPI ($0.02$) & 79.90 & 79.40 & 65.70 \\
CDPI ($0.04$) & 77.20 & 78.28 & 64.63 \\
\midrule
Rnd. seed 11 & 79.40 & 78.12 & 64.75 \\
Rnd. seed 42 & 78.60 & 78.64 & 65.59 \\
Rnd. seed 67 & 77.00 & 78.32 & 64.52 \\
Rnd. seed 527 & 79.20 & 79.48 & 65.02 \\
Rnd. seed 1483 & 78.00 & 78.48 & 64.65 \\
\midrule
Rnd. $\mu\!\pm\!\sigma$ & $78.44\!\pm\!0.97$ & $78.61\!\pm\!0.52$ & $64.91\!\pm\!0.42$ \\
\bottomrule
\end{tabular}
\caption{Random-control perception results (part 2 of 2). Perception averages all five benchmarks.}
\label{tab:supp_random_perception_cont}
\end{table}

\FloatBarrier

The random perturbations do not reproduce the directional transfer pattern.
Their high-level reasoning average is
$24.02\pm1.31$, below both the target score of $24.96$ and the CDPI score of
$26.66$ at $\alpha=0.02$. The random control also decreases the General and
Perception averages relative to the target. At the benchmark level, it does
not reproduce the CDPI improvements on MATH-Vision, VisuLogic, or
VisualPuzzles.

\bibliography{references}


\end{document}